\documentclass{article}

\usepackage[preprint,nonatbib]{modified_neurips_2023}

\usepackage{amsmath, amssymb, amsfonts}
\usepackage{bbm}
\usepackage{dsfont}
\usepackage{xcolor}
\usepackage[bottom]{footmisc}
\usepackage{tabularx}
\usepackage{tablefootnote}
\usepackage[verbose=true,letterpaper,tmargin=1in,bmargin=1in,left=1.15in,right=1.15in]{geometry}
\usepackage[breaklinks]{hyperref}   %
\usepackage[noabbrev,capitalize]{cleveref}
\usepackage[bottom]{footmisc}
\usepackage{graphicx}
\usepackage{caption}
\usepackage{subcaption}
\usepackage{xspace}
\usepackage[normalem]{ulem}
\usepackage{multirow}
\usepackage{float}
\usepackage{ulem}
\usepackage{multicol}
\usepackage{enumitem}
\usepackage{cuted}
\usepackage{url}
\usepackage{setspace}
\usepackage{fancyhdr}
\usepackage{CJKutf8}

\crefformat{footnote}{#2\footnotemark[#1]#3}
\def\eqref#1{equation~\ref{#1}}

\def\1{\bm{1}}

\DeclareMathAlphabet{\mathsfit}{\encodingdefault}{\sfdefault}{m}{sl}
\SetMathAlphabet{\mathsfit}{bold}{\encodingdefault}{\sfdefault}{bx}{n}

\newcommand{\ifprecedingtext}[1]{\ifvmode\relax\else#1\fi}

\newenvironment{blue}{
    \color{blue}
}{
    \ignorespacesafterend
}
\newenvironment{orange}{
    \color{orange}
}{
    \ignorespacesafterend
}
\newenvironment{olive}{
    \color{olive}
}{
    \ignorespacesafterend
}

\usepackage[backend=biber,style=nature,natbib=true]{biblatex}
\bibliography{main}

\title{Disentangling concept semantics via multilingual averaging in Sparse Autoencoders}

\author{%
  Cliff O'Reilly \\
  Department of Computer Science \\
  City St George's, University of London \\
  \texttt{cliff.oreilly@citystgeorges.ac.uk} \\
  \And
  Ernesto Jim\'{e}nez-Ruiz  \\
  Department of Computer Science \\
  City St George's, University of London \\
  \texttt{ernesto.jimenez-ruiz@citystgeorges.ac.uk}
  \And
  Tillman Weyde  \\
  Department of Computer Science \\
  City St George's, University of London \\
  \texttt{t.e.weyde@citystgeorges.ac.uk}
}

\begin{document}

\maketitle

\begin{abstract}
Connecting LLMs with formal knowledge representation and reasoning is a promising approach to address their shortcomings. 
Embeddings and sparse autoencoders are widely used to represent textual content, but the semantics are entangled with syntactic and language-specific information. 
We propose a method that isolates concept semantics in Large Langue Models by averaging concept activations derived via Sparse Autoencoders. 
We create English text representations from OWL ontology classes, translate the English into French and Chinese and then pass these texts as prompts to the Gemma 2B LLM. 
Using the open source Gemma Scope suite of Sparse Autoencoders, we obtain concept activations for each class and language version. 
We average the different language activations to derive a \textit{conceptual average}. 
We then correlate the conceptual averages with a ground truth mapping between ontology classes.
Our results give a strong indication that the conceptual average aligns to the true relationship between classes when compared with a single language by itself. 
The result hints at a new technique which enables mechanistic interpretation of internal network states with higher accuracy. 
\end{abstract}

\section{Introduction}

The combination of LLMs with formal reasoning and knowledge representation has  become a topic of increased interest recently. 
While the improvements of LLMs have enabled new applications of machine learning and artificial intelligence, LLMs have weaknesses, such as hallucinations and reasoning errors. 
The combination of LLMs with formal knowledge representation, such as ontologies, has the potential to address these problems but it requires bridging the gap between text and formal semantics. 
Sparse autoencoders can help model semantic content in an interpretable way by disentangling components in text embeddings. 
However, embeddings contain not only information on concept semantics, but also syntactic and language-specific aspects. 
In this work, we propose the use of multi-lingual aggregation to extract concept semantics and evaluate this method in ontology alignment tasks. 

Our question is: can multi-lingual representations be used to isolate semantic aspects in sparse embedding vectors. 
This will help both mechanistic interpretability, but also practical tasks like ontology alignment, and more generally the integration between LLMs and formal sematic systems. 

In our method, we parse OWL ontology classes into text representations and use these as prompts for an LLM with Sparse Autoencoders (SAEs), which produces concept activations. 
We use existing class similarity mapping from the ontology suite\footnote{\url{https://oaei.ontologymatching.org/2024/conference/index.html}} as ground truth, and compare the similarity of concept activations  with the ground truth. 
With single language prompts in English, concept  activations are noisy, in that they show low correlation to the ground truth. 
To address this, we perform a natural language translation to produce multiple concept activation vectors. 
The average of these vectors (which we term the \textit{conceptual average}) shows a clearly stronger correlation to ground truth.  
We interpret this as evidence for syntactic and language-specific information being suppressed by the multi-lingual approach. 

The rest of this paper is organised as follows: 
In \autoref{sec:related} we present the context of our idea and related work, in \autoref{sec:method}, we introduce our method followed by results, in \autoref{sec:results}, and a discussion in \autoref{sec:discussion}.

\section{Related Work}\label{sec:related}

An \textit{ontology} in general terms can be thought of as a formal representation of a domain of knowledge. Any representation has to be subjective and, in practice, ontologies tend to be bespoke to a domain or application. 

\subsection{Ontology Parsing}

OWL\footnote{\url{https://www.w3.org/OWL/}} ontologies are a standard, machine-readable and flexible format for representing any domain. Due to the subjective nature of semantic representation, our goal of creating a text prompt from an OWL ontology is also subjective. The extraction of OWL classes, properties and relationships can be performed with libraries for various programming languages (we used OWLAPI \citep{horridge2011owl}), and tools have been created to generalise text extraction, e.g. NaturalOWL \citep{androutsopoulos2013generating} and OWL Verbalizer \citep{kaljurand2007verbalizing}. The \textit{recursive concept verbaliser} approach for ontology subsumption inference \citep{he2023language} presents a toolbox for OWL ontology analysis (OntoLAMA). 

\subsection{Ontology Alignment}

The challenge of matching concepts between ontology representations is as old as the representations themselves. Since 2004, the Ontology Alignment Evaluation Initiative\footnote{\url{https://oaei.ontologymatching.org/}} has provided a framework for evaluating various approaches. From straightforward lexical approaches, through structural and semantic techniques to more recent innovations with machine learning \citep{qiang2023agent} (and a multitude of hybrid methods (\citealp{euzenat2004ontology}; \citealp{codescu2014categorical}; \citealp{jimenez2011logmap})), we believe our research is novel in approaching the problem with analyses of LLM internal concept states.

\subsection{Mechanistic Interpretability and Sparse Autoencoders}

Mechanistic Interpretability (MI) is a domain which aims to interpret the internal activation states of neural networks for various purposes such as AI safety, neural network decision-making and improving network design \citep{sharkey2025open}. Our interest in MI is for the learned concept activations --- the correlation between conceptual semantics and node activations of LLMs. 

When applied to language models, Sparse Autoencoders are unsupervised algorithms that learn to map from latent representations to interpretable concepts (also called features). An SAE is a pair of encoder and decoder functions that compresses an input into a hidden representation and tries to reconstruct the input from the hidden representation --- thereby learning a set of activation features which can be correlated via techniques such as Dictionary Learning \citep{bricken2023towards} to a vocabulary of human understandable concepts. The sparsity controls that are applied during training result in a reduced set of activations that are more easily computed (compared with billions of activations in a full LLM).

Gemma Scope \citep{lieberum2024gemma} is an open suite of SAEs trained on Google's Gemma 2 LLM --- at every layer and sublayer.

\section{Method}\label{sec:method}

The corpus used in this experiment comes from the conference track of the Ontology Alignment Evaluation Initiative 2024. Across the 16 ontologies, there are 867 class definitions and a set of 174 reference class mappings as ground truth alignments. The alignment mappings do not cover all the 16 ontologies so we cut down the analysis to only those ontologies with a target mapping available.

The method we apply can be broken down into stages, described below.

1. We use the Java library \textit{OWLAPI}\footnote{\url{https://github.com/owlcs/owlapi}} to parse each owl ontology file. Due to the nature of owl representations, each ontology can take a different format and so the Java script has bespoke logic that extracts classes, any related subclasses, superclasses, object properties and data properties. Some manipulation of the representation is needed. There are two styles of output we create: a summary and a verbose version --- summary and verbose examples for the class Author (from the \textit{edas} ontology) are shown in quotes (A) and (B), below. The verbose output is a text string which encapsulates a description of the class and includes connecting and descriptive words, but the summary version is a concatenation of the target class name, associated class names and object properties. 

\begin{quote}
(A) Author is a SuperClassOf Presenter and hasRelatedPaper Paper
\end{quote}

\begin{quote}
(B) Author is a SubClassOf some writes Contribution and is a SubClassOf Person and is a SubClassOf only writes Contribution and writes Contribution 
\end{quote}

2. We use the \textit{googletrans} Python library\footnote{\url{https://pypi.org/project/googletrans/}} to perform a natural language translation from English to French and English to Simplified Chinese. There are no parameters supplied to this process - it is a straightforward translation service. Examples of French and Chinese translations of summary and verbose representations are shown below:

\begin{quote}
(C) Personne auteur uniquement Contribution Certaines écritures Contribution Contribution
\end{quote}

\begin{quote}
(D) L'auteur est une sous-classe de contribution des écritures et est une personne sous-classe et est une sous-classe unique en rédaction de contribution et écrit la contribution
\end{quote}

\begin{quote}
(E) \begin{CJK*}{UTF8}{gbsn}
作者有些人写贡献只写贡献
\end{CJK*}
\end{quote}

\begin{quote}
(F) \begin{CJK*}{UTF8}{gbsn}
作者是一个子类人，是一个仅写作贡献的子阶级，并且是某些撰写贡献的子类别，并写下了贡献
\end{CJK*}
\end{quote}

3. Using the \textit{huggingface} library\footnote{\url{https://huggingface.co/google/gemma-scope}} \citep{wolf-etal-2020-transformers} to access the Gemma Scope open suite of sparse autoencoders, we process each text representation as a prompt to a PyTorch neural network (using a Jump ReLu activation function). The particular SAE set used here is the 2 billion parameter model based on Google's Gemma 2 Large Language Model. We take every layer (0 to 25) of the 16.4k width model and we take the L0 Norm variant for model regularisation (the number of non-zero elements in the activation vector) where the average is between 13 and 23 active features (e.g. 13 out of 16.4k on average). If L0 is set too high then features overlap and interpretability breaks down. Set L0 too low and the network underfits and misses important structures. The output is a set of concepts and an activation weighting. An example tensor with 7 concept identifiers and activation values is shown in Table \ref{tab:extensor1}.

\begin{table}[H]
  \centering
  \begin{tabular}{lll}
    \hline
    \textbf{Concept ID} & \textbf{Activation Weight} \\
    \hline
2446 & 57.0846 \\
3391 & 47.3293 \\
3752 & 37.2378 \\
5327 & 79.6517 \\
6035 & 70.1694 \\
6035 & 71.6481 \\
7234 & 36.0779 \\
8816 & 46.0310 \\
9823 & 57.1111 \\
10144 & 43.9628 \\
12529 & 49.0565 \\
14829 & 61.3937 \\
    \hline
  \end{tabular}
 
  \caption{Example concept activation set 1 (for the verbose English edas-Author class)}
  \label{tab:extensor1}
\end{table}

4. The same process as in (3) is repeated for the translated (French and Chinese) texts.

\begin{figure}[H]
    \centering
    \includegraphics[width=0.9\linewidth]{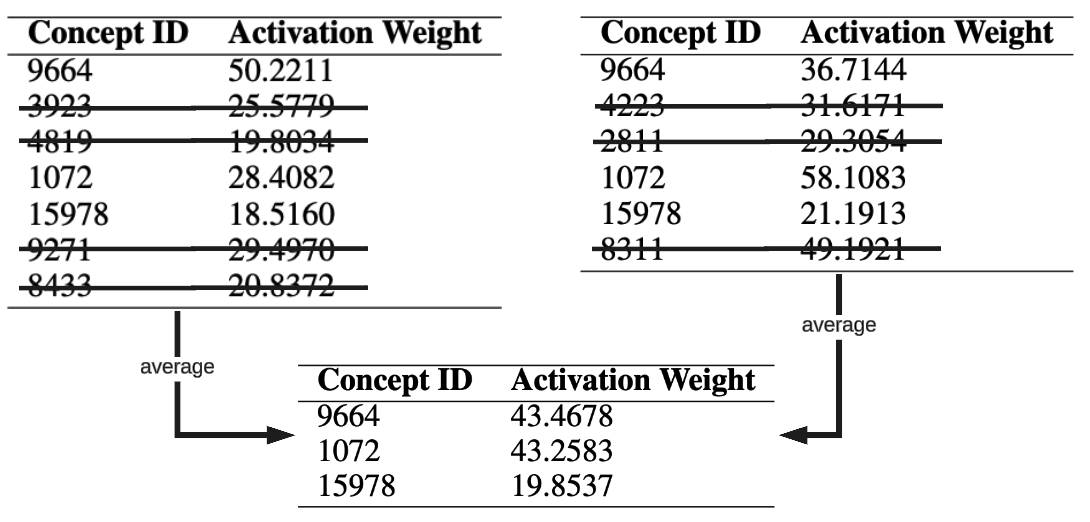}
    \caption{Converting two concept sets into a conceptual average}
    \label{fig:conceptualaverage}
\end{figure}

5. For each class representation, the English and the translated concept activation sets are averaged. For concepts shared between the English and translated sets the simple average of the weighting is taken. Concepts not shared are removed from the output. This results in a much reduced set of concepts for each class. The resulting average class representation we call the \textit{conceptual average}. An example is shown in Figure \ref{fig:conceptualaverage}.

\begin{figure}[H]
    \centering
    \includegraphics[width=0.9\linewidth]{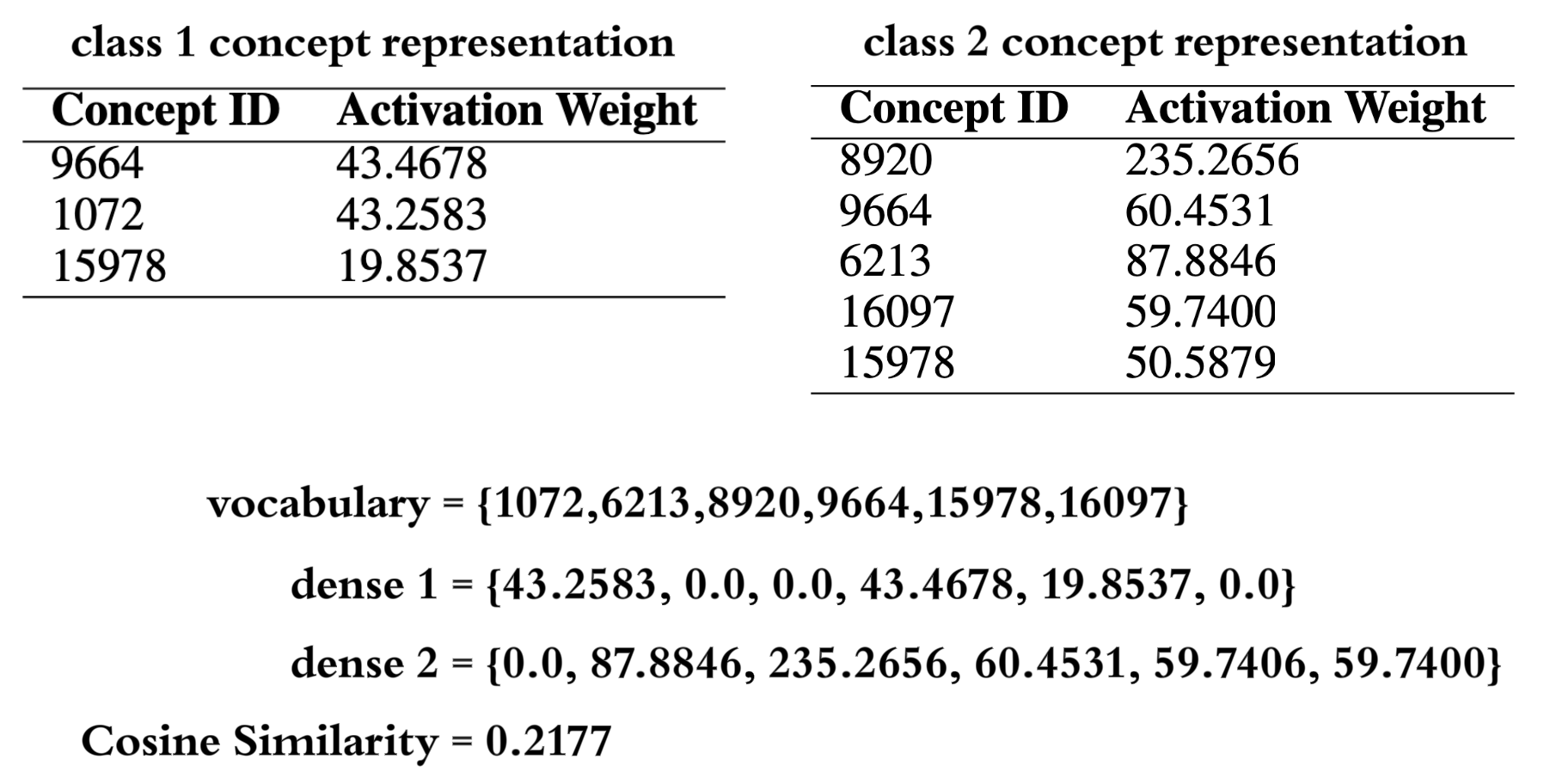}
    \caption{Computing the class concept representation similarity}
    \label{fig:computresim}
\end{figure}

6. Every class average within each of the three groups (English only, English to French combined conceptual average and English to Chinese conceptual average) is compared with every other class average in the same group using a Cosine Similarity function (function represented in Figure \ref{fig:computresim}. Where the class comparison has a pre-defined mapping in the ground truth dataset, we align the similarity score with a target variable value of \textit{1}, else it is set to \textit{0}. An example record, showing the similarity score of the \textit{Author} classes from the \textit{emt} and \textit{edas} ontologies, with a target of \textit{1} is shown below, at (H).

\begin{quote}
(H) \textit{cmt-Author,edas-Author,0.8362799,1}
\end{quote}

7. The resulting output from the previous steps is a set of differences between representations of classes from the source ontologies. The ground truth class reference mappings are used to create a correlation between the correct relationship and the conceptual difference. The correlation is a measure of the accuracy of conceptual representation from the sparse autoencoder, and by creating the conceptual average we hope to improve the accuracy. The correlation algorithm used is the Point-Biserial Correlation, which is ideal for  correlations between binary and continuous variables.

Due to the nature of the corpus, the layer-by-layer analysis only has 174 ground truth mappings (from a total of 95,000 class comparisons) and hence there is a large class imbalance for each layer. We reduce the imbalance by using a random re-sampling to reduce the \textit{false} target variable size to be the same as the \textit{true} size.

A further analysis was undertaken using the same approach, but instead of translating to French, Simplified Chinese was used.

Example code is available for validation\footnote{\url{https://github.com/cliffore/millms}}.

\section{Results}\label{sec:results} 

The corpus used in this experiment contains 16 OWL ontology files, comprising 867 classes. The corpus also includes 174 class mappings which form the ground truth of this experiment. 

From this set of files, we follow the method described above and record the results as shown below.

The outputs consist of a correlation between ontology classes and over a number of variations such as the nature of the text representation (either a summary or verbose representation) and the layer of concept activation (0-25). These results are then compared at the level of language, i.e. are the English-only concept sets different from the French/Chinese conceptual averages.

\begin{figure}[H]
    \centering
    \includegraphics[width=1\linewidth]{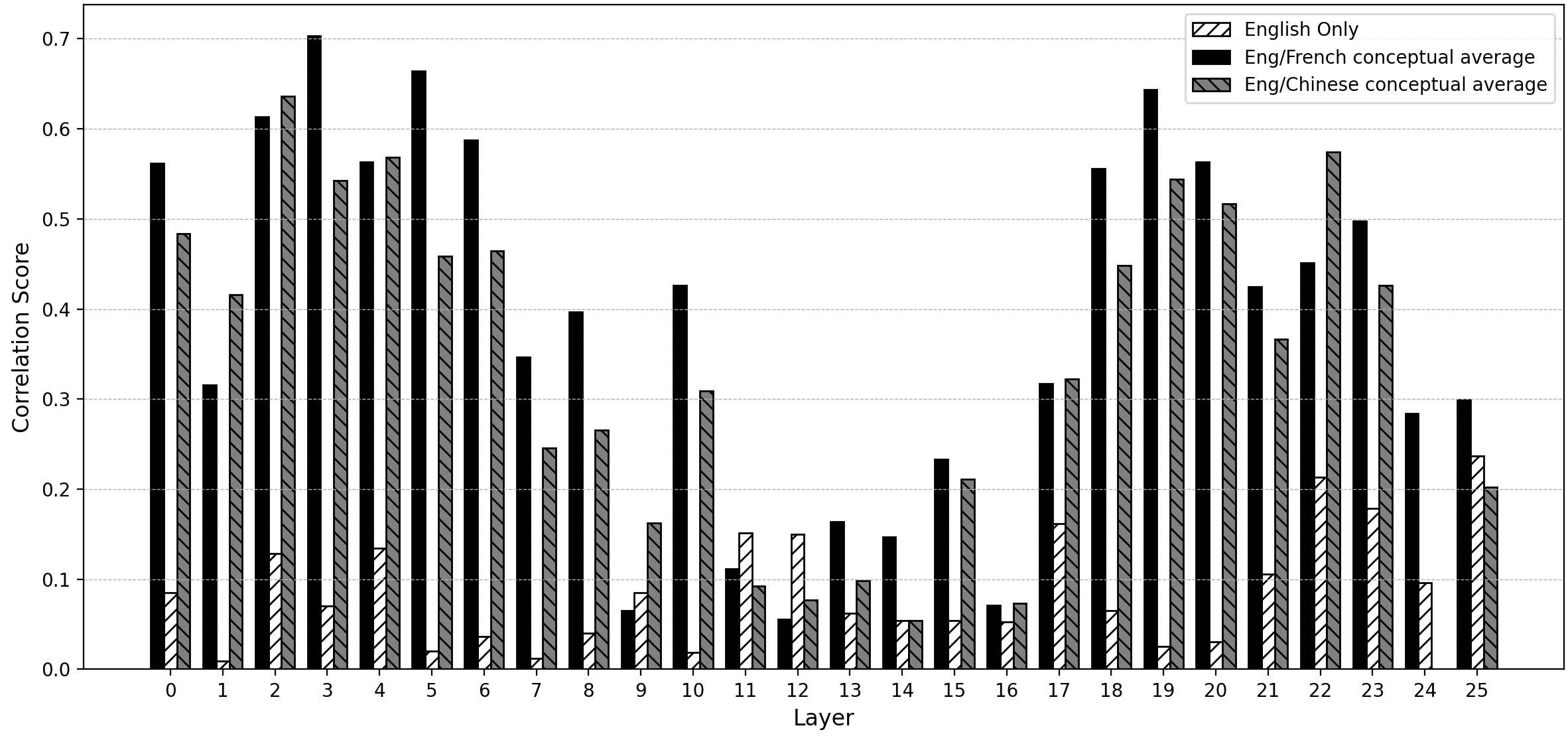}
    \caption{Summary prompt - conceptual average correlation vs English-only (higher is better)}
    \label{fig:correlations}
\end{figure}

\begin{figure}[H]
    \centering
    \includegraphics[width=1\linewidth]{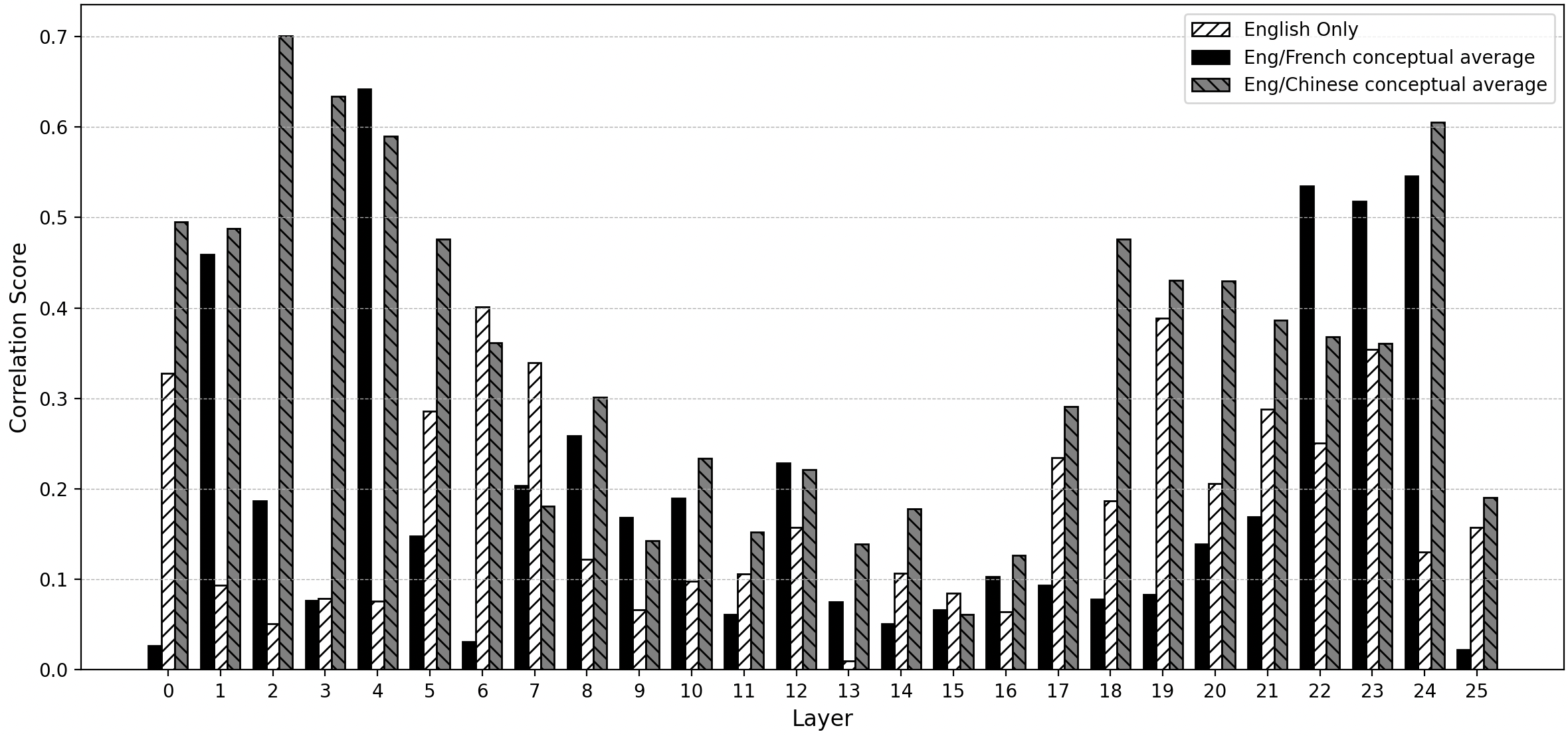}
    \caption{Verbose prompt - conceptual average correlation vs English-only (higher is better)}
    \label{fig:correlationv}
\end{figure}

Results are shown in Figure \ref{fig:correlations}, for the summary text representation, and Figure \ref{fig:correlationv} for the verbose prompt. Both results show a clear improvement in correlation comparing English only and the translation conceptual average. There is a small increase in correlation when we look at averages of translations between English and Simplified Chinese for summary texts, but a reduction for the verbose texts.

Table~\ref{tab:results1} shows the correlations compared.

\begin{table}[H]
  \centering
  \begin{tabular}{lll}
    \hline
    \textbf{Text Version} & \textbf{Language} & \textbf{Correlation}  \\
    \hline
    Summary & English only & 0.09  \\
    Summary & Avg Eng/French & 0.39  \\
    Summary & Avg Eng/Chinese & 0.33 \\
    Verbose & English only & 0.18  \\
    Verbose & Avg Eng/French & 0.20  \\
    Verbose & Avg Eng/Chinese & 0.35  \\
    \hline
  \end{tabular}
 
  \caption{Average correlations for each text version and the translated language}
  \label{tab:results1}
\end{table}

\section{Discussion}\label{sec:discussion}

After we extract an English text version of an ontology class, we put that string through a set of SAE neural networks and compare the concept activations between other classes from related ontologies. We have a (small) set of ground truth correspondences between classes and we see a fairly weak correlation emerge. This is potentially due to the small size of the corpus and also the relatively subjective extraction from OWL representation to a string of words. We notice that there is a difference in overall correlations between the summary extracted text and the more verbose version.

We take the same English text and translate it to French and Simplified Chinese and put these two prompts through the same SAEs. The resulting concept activation sets are averaged for each ontology class between the English and French and between the English and Chinese versions. This represents a different concept activation set for each ontology class. When we calculate the same correlations as with the English only activations, we see a significant improvement in correspondence for the French translation and an even stronger correlation for the Chinese translations. The difference in average correlations between the French and Chinese (from Table \ref{tab:results1}) is small, however the average percentage difference is 9\% (summary).

In the same way that ``Conceptual Semantics takes the meanings of words and sentences to be structures in the minds of language users'' \citep{jackendoff2006conceptual}, we might assume that LLMs have structures which represent the meaning of words that are processed through their neural network states. We might also assume that LLMs don't have concepts of meaning in themselves, but instead are learning and storing correspondences between symbolic and linguistics structures upon which LLMs are trained. Arguments are emerging, however, which show that LLMs do represent real world concepts (\citealp {gurnee2023language} and \citealp{kim2025linear}) beyond the purely linguistic. When we peek into the network internals, using Sparse Autoencoders, we often see that concepts are activated which relate to surface cues such as syntactic and linguistic semantics, but our results show that we can reduce the symbolic concept space for a set of activations and isolate concepts that reflect a purer semantic representation.

Following the example from before, we had an \textit{Author} class that was represented as English, French and Chinese. The verbose representation concept activation set is shown in Table \ref{tab:extensor1}. In Table \ref{tab:extensor2}, we show the Chinese translation concept activations, and in Table \ref{tab:extensor3} the conceptual average output.

\begin{table}[H]
  \centering
  \begin{tabular}{lll}
    \hline
    \textbf{Concept ID} & \textbf{Activation Weight} \\
    \hline
2446 & 21.1420 \\
5327 & 44.4837 \\
6035 & 39.4718 \\
6035 & 39.7229 \\
7748 & 30.9887 \\
8920 & 22.0814 \\
9967 & 73.0786 \\
13833 & 15.4156 \\
14763 & 19.7685 \\
    \hline
  \end{tabular}
 
  \caption{Example concept activation set 1 (for the verbose Chinese translation of edas-Author class)}
  \label{tab:extensor2}
\end{table}

\begin{table}[H]
  \centering
  \begin{tabular}{lll}
    \hline
    \textbf{Concept ID} & \textbf{Activation Weight} \\
    \hline
2446 & 39.1133 \\
5327 & 62.0677] \\
6035 & 55.6855 \\
    \hline
  \end{tabular}
 
  \caption{Conceptual average activation set (for the verbose English and Chinese translation of edas-Author class)}
  \label{tab:extensor3}
\end{table}

Concepts can be interpreted and given human-readable representations \citep{paulo2024automatically}; concept  ID 3391 in this analysis can be interpreted as (G), below (more detail can be seen on the Neuronpedia website e.g. for concept 6035 on layer 0: \url{https://www.neuronpedia.org/gemma-2-2b/0-gemmascope-att-16k/6035}. This concept is removed during the averaging process for the \textit{edas-Author} class because it is only activated by the English text. The output conceptual average has interpretations shown in Table \ref{tab:exautointerp1} below.

\begin{quote}
(G) attends to key parameters denoted by brackets from associated values in the same context
\end{quote}

\begin{table}[H]
  \centering
  \begin{tabular}{lp{10cm}}
    \hline
    \textbf{Concept ID} & \textbf{Automatic interpretation} \\
    \hline
2446 & attends to the token "Sant" from related tokens concerning "orum" and "ana." \\
5327 & attends to numerical outputs and associated metadata from other \newline mathematical or technical symbols and structures \\
6035 & attends to specific instances of the token "N" from various indexed \newline digits appearing later in the sequence \\
    \hline
  \end{tabular}
 
  \caption{Conceptual average activation set (for the verbose English and Chinese translation of edas-Author class)}
  \label{tab:exautointerp1}
\end{table}

The interpretation representations don't intuitively correspond with a human-readable assessment of similarity or conceptual semantics, however we suggest that our results show some significance to the process of removing concepts that are not shared between translated representations. The translation and averaging process is removing linguistic specific concepts and leaving concepts that are a purer representation of the core semantics of the original prompt.

The overall average results (Table \ref{tab:results1} indicate a trend for the Chinese conceptual average to correlate more strongly with the ground truth when compared with the French translation (assuming that the Summary versions (0.39 and 0.33) are too similar to be divided). This pattern adds credence to the argument since Chinese language tokens contains fewer syntactic elements, ``e.g. more frequent functional words in English texts'' \citep{wang2024translation}. Given that the dataset is small and the representations and translations relatively subjective, this result should be validated.

This slightly unexpected result hints at a new technique for improving conceptual analyses of LLMs, especially via SAEs. We expect future research to confirm and extend this result.

\section{Future Work}

We highlight some problems which we hope to address in future versions of the research.

The accuracy of extraction of class representations is exposed to problems of subjectivity. Both the conceptual model used to create an OWL ontology and the extraction process to create a text string version are prone to idiosyncrasies in design.

The corpus used is relatively small, having a low number of OWL classes. There is also a class imbalance because there are missing ground truth mappings for many OWL classes. The ground truth is also liable to potential bias.

The use of SAEs for interpretability is a relatively novel approach and there are known challenges e.g. feature splitting \citep{chanin2024absorption}, terse concept dictionaries and reconstruction errors \citep{shu2025survey}.

The following areas are in scope for next steps: (\textit{i}) Extend the corpora to confirm and explore this result,
(\textit{ii}) Explore a generalised ontology class extraction process,
(\textit{iii}) Analyse concept features for a common sense analysis,
(\textit{iv}) Apply improvements in interpreting conceptual semantics to Ontology Alignment tasks.


\printbibliography

\end{document}